\documentclass{ifacconf}

\usepackage{graphicx}      % include this line if your document contains figures
\usepackage{natbib}        % required for bibliography
\usepackage{color}
\usepackage{amsmath}
\usepackage{booktabs, amsfonts, dcolumn}
\usepackage{multicol}
\usepackage{subfigure}
\usepackage{tikz}
\usepackage{algorithm,algpseudocode}
\usepackage{url}
\usepackage{mars}

\usetikzlibrary{positioning,fit,arrows.meta,backgrounds}

\tikzset{
    module/.style={%
        draw, rounded corners,
        minimum width=#1,
        text width = 20mm,
        align=center,
        fill={rgb:red, 1; green, 4; blue, 1},
        minimum height=7mm,
        font=\sffamily
        },
    module/.default=2cm,
    >=LaTeX
}

\begin{document}

%
% COPY FROM HERE
%

% MARS title page generation:

% If not submitted/ accepted anywhere, comment \marsPublishedIn and \marsConference
% Choose one:
% \marsPublishedIn{Submitted to:} 
% \marsPublishedIn{To be published in:}  	% for Journals
\marsPublishedIn{Accepted for:} 		% for Conferences
% \marsPublishedIn{Published in:} 		% for Journals
% \marsPublishedIn{Published with:} 	% for Conferences

% The name of the venue. Will also be used in \marsMakeCitation. Include the year for conferences. 
% Be sure to include IEEE if it is an IEEE conference (e.g. IEEE Conference on Robotics and Automation (ICRA) 2019 ). 
\marsVenue{IFAC World Congress 2020}

% Year of the publication (e.g. year of the conference, not submission) 
\marsYear{2020}

% You should replace the authors from above here, if they contain special marks for affiliation or so on. Otherwise comment.
\marsPlainAutors{Xiaoling Long, Qingwen Xu, Yijun Yuan, Zhenpeng He and S\"oren Schwertfeger}

% This generates a citation. The paper title without formatting goes in the first argument, the publisher, (for Journal also volume and issue number and page) in the second. Authors are either the ones given by \author or they are overwritten with \marsPlainAutors. Conference or Journal Name from above. Year from above, too.

\marsMakeCitation{Improved Visual-Inertial Localization for Low-cost Rescue Robots}{}

% Link to the DOI.
\marsDOI{\url{}}

% Use this if this is an IEEE publication. It will use the year from above.
%\marsIEEE{}

% Use this for a custom copyright text
%\marsCopyright{\copyright~\marsYearName~Springer. Bla }

% Actually make the MARS titlepage. Comment if you don't want to generate it.
%\makeMARStitle

\onecolumn

\includegraphics[width=0.12\linewidth]{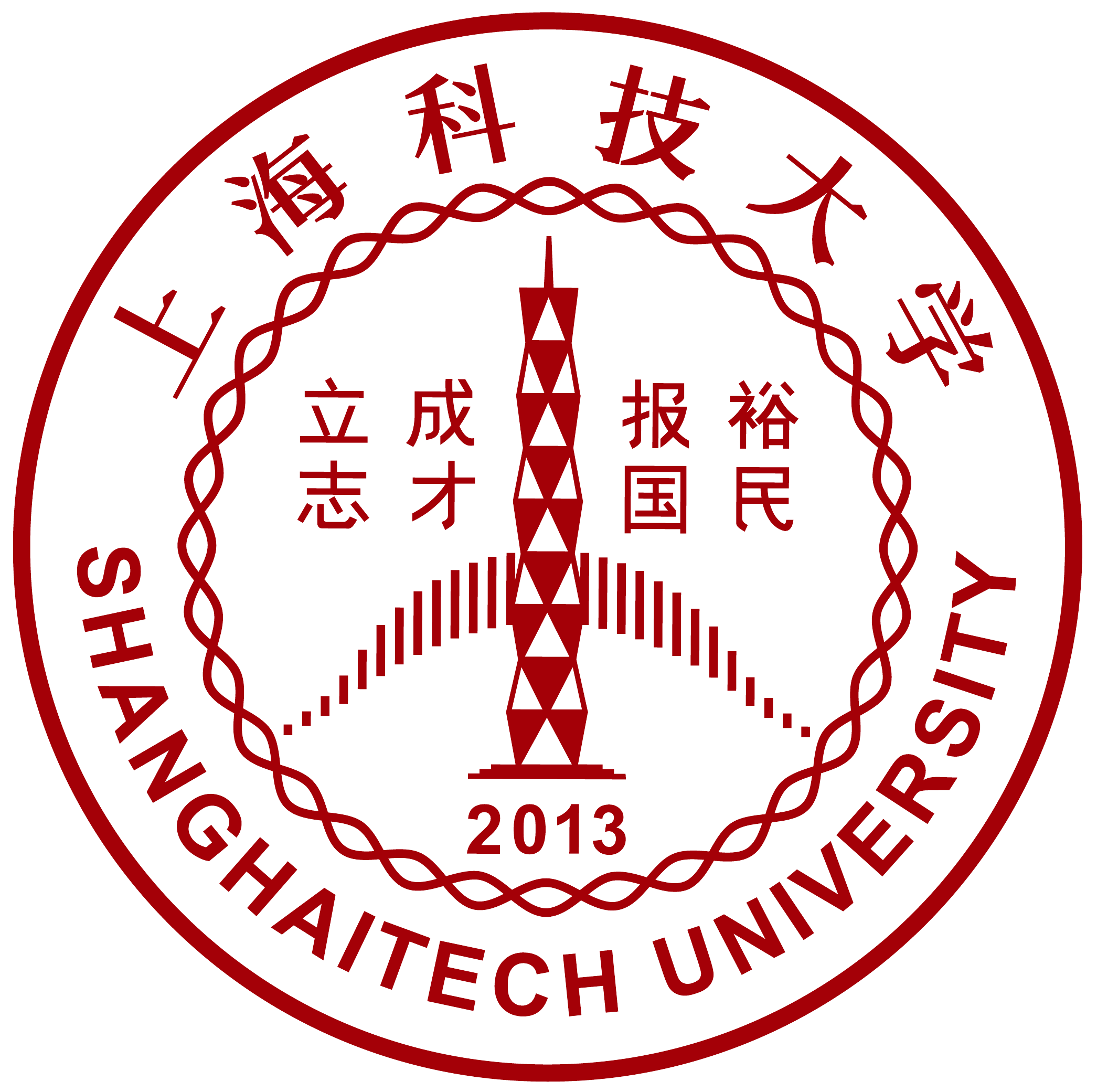}
\hspace{.02\linewidth}
\includegraphics[width=0.12\linewidth]{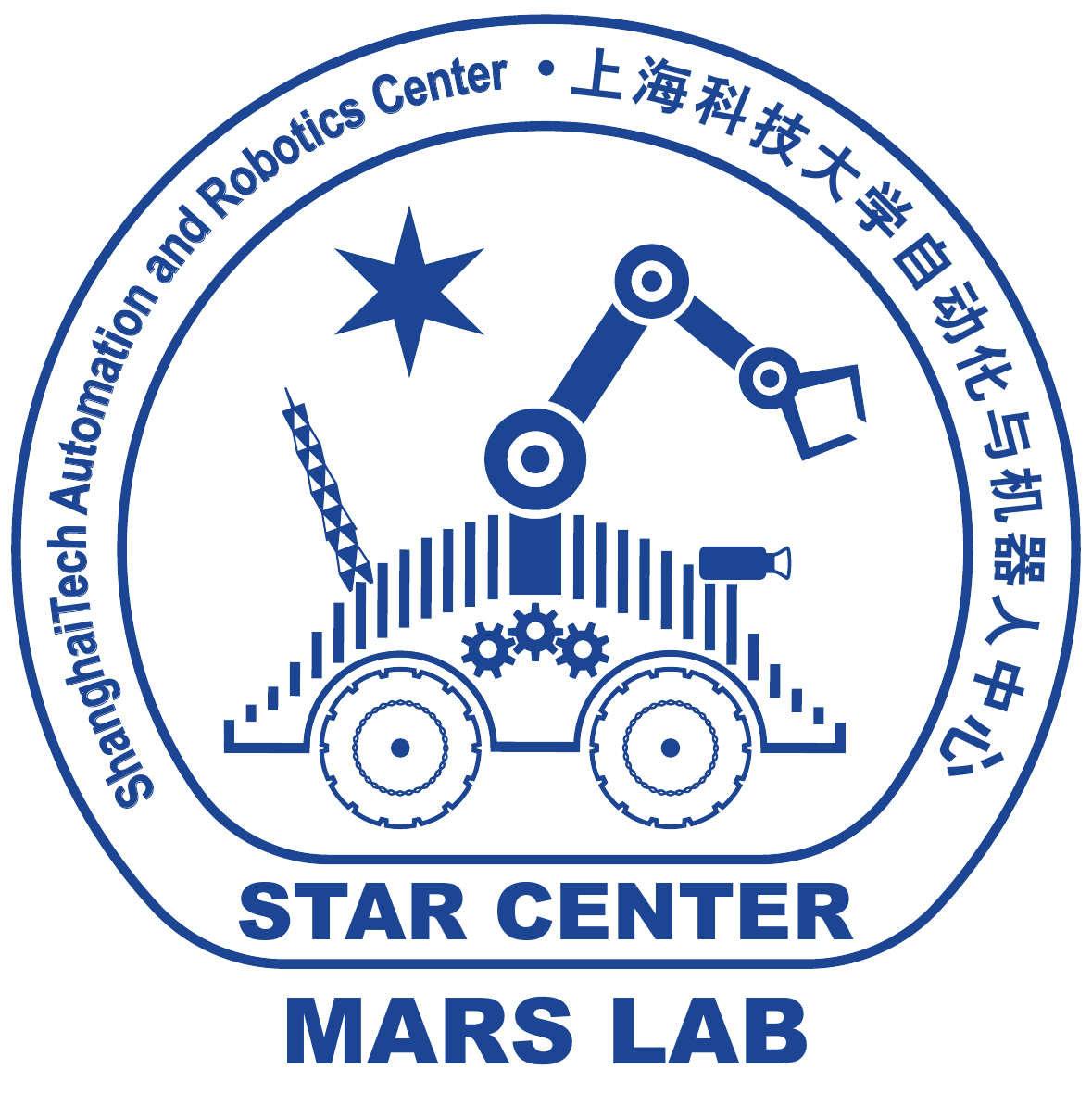}
\vskip.05in{ \color{marsBlue} \rule{\linewidth}{0.001pt} \\ }	

\begin{center}%

	\vspace{1cm}{\LARGE{Improved Visual-Inertial Localization \\ for Low-cost Rescue Robots}\par}\vskip1.0em\par%
	
	{ \color{marsBlue} \rule{0.8\linewidth}{1.4pt} \\ }

	\vskip0.65in{\Large\marsPublishedInName\par}\vskip.5em\par%
	{\LARGE\marsVenueName\par}\vskip1.0em\par%
	
	\vskip1.25in
	
\end{center}

 \ifx \marsCitationSet \undefined \else
{\Large Citation:}\\
{ \color{marsBlue} \rule{\linewidth}{0.4pt} \\ }
\vskip0.0em{\marsCitationName}\vskip0.45em\par%
\fi

{ \color{marsBlue} \rule{\linewidth}{0.4pt} \\ }

\ifx \marsDOIset \undefined \else
\vskip0.25in{\large DOI: {\marsDOIName }\par}\vskip.50em\par%
\fi

\vfill

This is a publication from the Mobile Autonomous Robotic Systems Lab (MARS Lab), School of Information Science and Technology (SIST) of ShanghaiTech University. For this and other publications from the MARS Lab please visit:\\
\url{https://robotics.shanghaitech.edu.cn/publications}

\vskip.50in\par

\marsCopyrightName

%
% END COPY
%

\begin{frontmatter}

\title{Improved Visual-Inertial Localization for Low-cost Rescue Robots }

\thanks[footnoteinfo]{The first two authors contributed equally.}

\author[First]{Xiaoling Long\thanksref{footnoteinfo}}
\author[First,Second]{Qingwen Xu\thanksref{footnoteinfo}}
\author[First]{Yijun Yuan}
\author[First]{Zhenpeng He}
\author[First]{S\"oren Schwertfeger}

\address[First]{School of Information Science and Technology, ShanghaiTech,
   China (e-mail: \{longxl, xuqw, yuanwj, hezhp, soerensch\}@shanghaitech.edu.cn).}
\address[Second]{Chinese Academy of Sciences, Shanghai Institute of Microsyst \& Information Technology, ShanghaiTech,China; University of Chinese Academy of Sciences, Beijing 100049, China}

\begin{abstract}                % Abstract of not more than 250 words.

This paper improves visual-inertial systems to boost the localization accuracy for low-cost rescue robots. When robots traverse on rugged terrain, the performance of pose estimation suffers from big noise on the measurements of the inertial sensors due to ground contact forces, especially for low-cost sensors. Therefore, we propose \textit{Threshold}-based and \textit{Dynamic Time Warping}-based methods to detect abnormal measurements and mitigate such faults. The two methods are embedded into the popular VINS-Mono system to evaluate their performance. Experiments are performed on simulation and real robot data, which show that both methods increase the pose estimation accuracy. Moreover, the \textit{Threshold}-based method performs better when the noise is small and the \textit{Dynamic Time Warping}-based one shows greater potential on large noise.

\end{abstract}

\begin{keyword}
Robotics, Fault Detection, State Estimation, Inertial Localization, Rescue Robot
% Five to ten keywords, preferably chosen from the IFAC keyword list.
\end{keyword}

\end{frontmatter}
%===============================================================================

\section{Introduction}

Visual-inertial systems have achieved great success during the last decades, both on methods and applications. \cite{huang2019visual} reviewed all kinds of visual-inertial approaches for state estimation, with vision from direct \citep{usenko2016direct} to feature-based \citep{mur2017visual} methods and fusion methods from filtering \citep{mourikis2007multi,li2012improving} to optimization \citep{indelman2013information,qin2017vins}. Also, \cite{kuang2019pose,xu2019improved} applied spectral methods for visual odometry. In addition, there are also multiple different applications: augmentation reality for mobile device like Google Tango\footnote{\url{https://www.google.com/atap/projecttango.}}; visual-inertial odometry on unmanned air vehicles \citep{ling2016aggressive,sun2018robust,do2019high}; visual-inertial systems on wheeled robots \citep{wu2017vins} and tracked robots \citep{shan2019rgbd}.

Assuming that the camera can provide reliable visual information, the inertial measurement unit (IMU) largely determines the performance of visual-inertial localization. The IMU is used to obtain the acceleration and angular velocity of the system, which can be integrated to calculate rough poses. However, it can also be the drift source of inertial-based localization.
Often visual-intertial systems are applied to flying robots or robots using wheels on smooth terrain and thus the reported performance of such system is often good. But, as mentioned in \citep{shan2019rgbd}, the accelerometer equipped on a rescue robot is susceptible to a lot of vibration and ground contact forces when the robot traverses on rugged terrain, which usually causes unreliable localization. For example, robots might fall down a step, so impulses appear on acceleration measurements. After the double integration on acceleration to get the pose, the noise enlarges substantially. Thus the inertial-based localization fails. 

In this work, we attempt to address such fault problems with the application on rescue robots. In other words, we focus on building robust and reliable visual-inertial systems for robots operating on rough terrain. The reason why we choose rescue robots as our main research object is, that such fault conditions of IMU is more likely to happen here than other robots which move on 2D planes, due to the challenging terrains.

%In rescue field, the small rescue robot plays more and more import role, benefit from light and flexibility. Small robot is able to search rescue area where big and heavy robot might crash building residue, causing unpredictable loss. In rugged terrain, due to the structural defect of wheel robot, tracked robot show more effieciency and ability. \\
%Self-localization is a extremely central technology in robotics research, which is same as rescue robotics. Accurate self-localization speed up the rescue operation. The IMU-based localization method provides ??. The IMU is used to obtain the acceleration and angular velocity of robot, which can be integral to robot pose. This is also the drift source of IMU-based localization. Combined with wheel odometry or visual information, the drift can be reduced significantly. \cite{qin2017vins} developed VINS-Mono show the big success of visual-inertial based localization.

%TODO related work. include some localization paper and vio stuff.
%Also IMU-based localization achieve the great success in robotic self-localization. However, in rugged rescue scenario, robot might fall down from swell(high place), which impulses appear on acceleration measurements. After twice integral on acceleration, the noise will be enlarge many times. Finally the IMU-based localization will fail.

\begin{figure}[tb]
\centering
\includegraphics[width=0.65\linewidth]{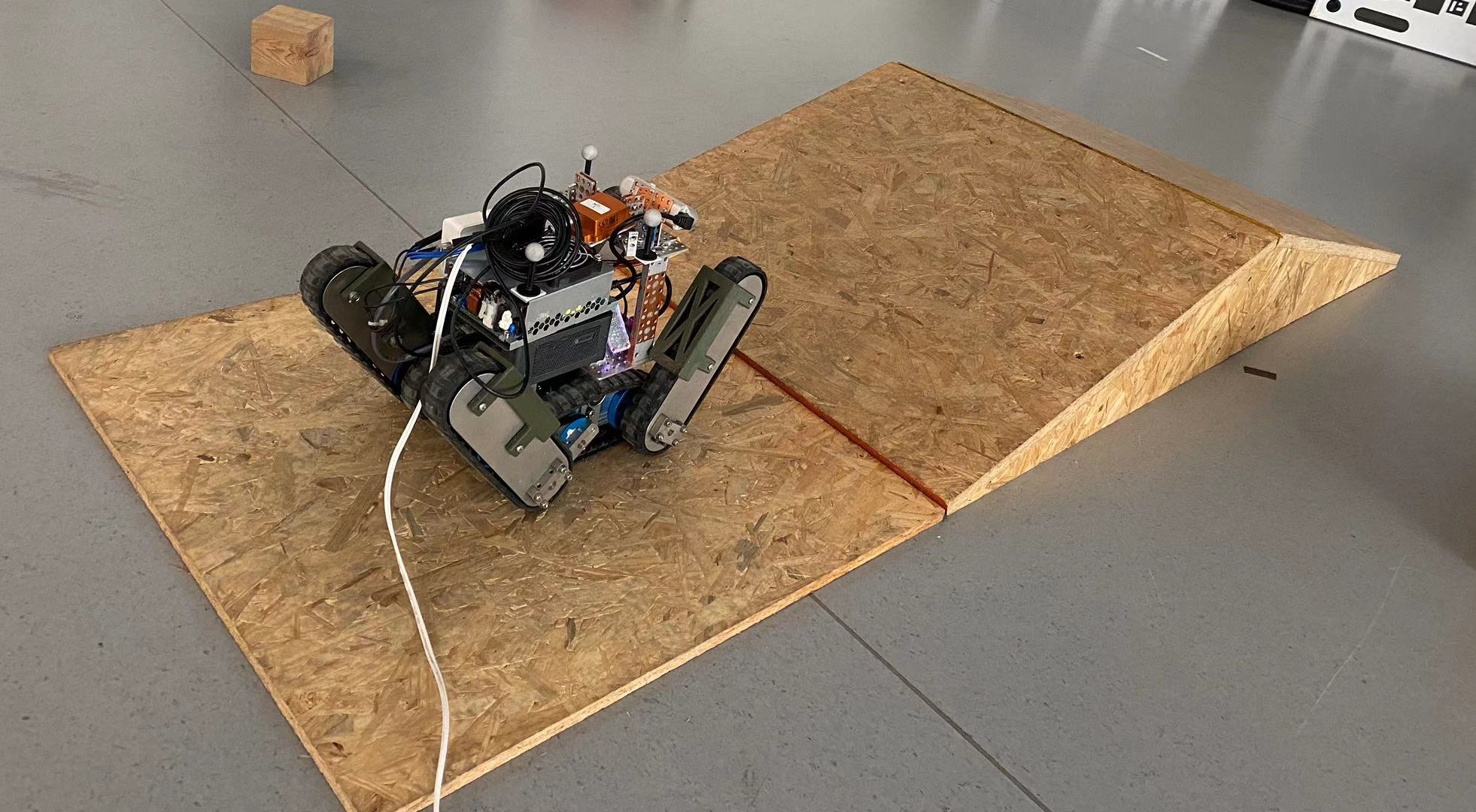}
\caption{MARS Rescue Robot used for the experiments.}
\label{fig:robot_setup}
\end{figure}

There have been some attempts already to try to deal with the instabilities of the IMU resulting from different causes, such as large noise, disturbance and disconnection. \cite{avram2015imu} detected the IMU bias fault based on the known roll and pitch for a UAV.
\cite{quan2019tightly} fuses vision, wheel encoders and gyroscope tightly to increase the accuracy and robustness of localization. \cite{nguyen2009diagnosis} proposed two model-based fault detection and isolation methods to maintain the robustness of a quadrotor, including parameter estimation and set membership estimation. In addition, multi-IMUs are also used to detect and isolate anomalies happened from each sensor. For example, \cite{d2017fault} proposed a filtering-based approach to detect faults on one of its two IMUs; \cite{eckenhoff2019sensor} utilized even more IMUs to address IMU failures.

The paper proposes \textit{Threshold}-based and \textit{Dynamic Time Warping} (DTW~\citep{berndt1994using})-based IMU preprocessing methods, which make inertial based localization more reliable and robust for low-cost robots. We summarize our contributions as:
\begin{enumerate}
   \item Detect IMU abnormal measurement values by both \textit{Threshold}-based and DTW-based methods. Further replace the measurement with reasonable data.
   \item Apply VINS-Mono on the processed data to illustrate the effectiveness of the proposed methods.
   \item Make a rescue robot have more reliable self-localization to achieve better performance.
\end{enumerate}

% In the following, we analyze the problem and introduce the proposed methods. After that,  simulation and real robot  experiments are utilized to evaluate our methods.
The rest of this paper is organized as follows: Sec.~\ref{sec:problem_analysis} analyzes and states the problem of IMU measurements. Then we propose the \textit{Threshold}-based and DTW-based methods to address this problem in Sec.~\ref{sec:method}. Afterwards, experiments are performed on both simulation data and real robot data to evaluate the proposed methods in Sec.~\ref{sec:exp}. Finally, we conclude our work in Sec.~\ref{sec:conclusion}. 

\section{Problem Analysis}
\label{sec:problem_analysis}

In this work, we diagnose the issue that the localization of our rescue robot equipped with the low-cost RealSense (see Fig.~\ref{fig:robot_setup}) often fails on rugged terrains when using the popular \underline{VI}sual  \underline{I}nertial  \underline{S}ystem  VINS-Mono (\cite{qin2017vins}). For that purpose, we firstly review the IMU model used in this system. Then the real IMU data collected from the rescue robot is used to analyze the localization failure problem.

%YIJUN: maybe with a ruler besides the robot?
%Xiaoling: No, we do not emphase "small"

\begin{table}[h]
\centering
\caption{Notation}
\begin{tabular}{l||l}
\hline\hline
%$\mathbf{a}$         & Acceleration               \\
%\hline
$\mathbf{a}^{w}$         & The acceleration at \emph{world} frame               \\
\hline
$\mathbf{a}^{b}$         & The acceleration at \emph{body} frame            \\
\hline
$\tilde{\mathbf{a}}^{b}$     &   The measurement of acceleration at \emph{body} frame     \\
\hline
$\mathbf{v}^{w}$         & The velocity at \emph{world} frame               \\
\hline
%$\mathbf{w}$         & The angular velocity               \\
%\hline
$\mathbf{w}^{b}$     & The angular velocity at \emph{body} frame \\
\hline
$\tilde{\mathbf{w}}^{b}$    &  The measurement of angular velocity at \emph{body} frame  \\
\hline
$\mathbf{n}$         & The white noise,  $ n \sim \mathcal{N}(0, \sigma^{2})$                                    \\
\hline
$\mathbf{n}^{g}$     & The white noise at \emph{gyroscopes} frame           \\
\hline
%$\mathbf{b}$         & The measurement bias                 \\
%\hline
$\mathbf{b}^{a}$     & The measurement bias at \emph{accelerometer} frame   \\
\hline
$\mathbf{b}^{g}$     & The measurement bias at \emph{gyroscope} frame   \\
\hline
$\mathbf{g}^{w}$     & The gravity at \emph{world} frame                 \\
\hline
$\mathbf{q}^{b}_{w}$ & Rotation from \emph{world} to \emph{body} frame in quaternion   \\
\hline
$\mathbf{p}_i, \mathbf{q}_i$ & The robot pose at time $i$ \\
\hline
$\mathbf{p}^{w}_i, \mathbf{q}^{w}_i$ & The robot pose at time $i$ at  \emph{world} frame \\
\hline
$d$ & The dimensionality of a single IMU measurement  \\
\hline
$\mathbf{P}, \mathbf{Q}$ & Two time series of IMU measurements   \\
\hline
$\mathbf{P}$ & One of the template patterns to be matched to   \\
\hline
$\mathbf{Q}$ & The actual IMU measurements   \\
\hline
$N, M$ & The length of  $\mathbf{P}, \mathbf{Q}$  \\
\hline
$k$ & The number of template patterns  \\
\hline
\end{tabular}
\end{table}

\subsection{IMU Model}

Considering the white noise and the bias from the random walk and ignoring the effect of scale, we can get the IMU measurement model as follows,
\begin{equation}
   \widetilde{\mathbf{w}}^{b} = \mathbf{w}^{b} + \mathbf{b}^{g} + \mathbf{n}^{g}
\end{equation}
\begin{equation}
   \widetilde{\mathbf{a}}^{b} = \mathbf{q}^{b}_{\mathbf{w}}(\mathbf{a}^{w} + \mathbf{g}^{w}) + \mathbf{b}^{a} + \mathbf{n}^{a}
\end{equation}
Since we have the derivative of position, velocity and quaternion at time $t$,
\begin{subequations}
   \begin{equation}
      \dot{\mathbf{p}}^{w}(t) = \mathbf{v}^{w}(t)
   \end{equation}
   \begin{equation}
      \dot{\mathbf{v}}^{w}(t) = \mathbf{a}^{w}(t)
   \end{equation}
   \begin{equation}
      \dot{\mathbf{q}}^{w}_{b}(t) = \mathbf{q}^{w}_{b}(t)\otimes
         \left[ \begin{array}{c}
            0 \\
            \frac{1}{2}\mathbf{w}^{b}(t)
         \end{array}\right]
   \end{equation}
\end{subequations}
 the pose of robot at time $t_{j}$ is obtained by integration during time interval $\left[t_{i}, t_{j}\right]$ :
\begin{subequations}
   \begin{equation}
      \begin{array}{l}
         \mathbf{p}^{w}(t_{j}) = \mathbf{p}^{w}(t_{i}) + \int_{t_{i}}^{t_{j}} \mathbf{v}_{t}^{w}(t)\delta t \\
            \quad\quad\quad   + \int\int_{t\in [t_{i}, t_{j}]} (\mathbf{q}^{w}_{b}(t)\mathbf{a}^{b}(t) - \mathbf{g}^{w})\delta t^{2}
      \end{array}
      % p^{w}(t_{j}) = p^{w}(t_{i}) + \int_{t_{i}}^{t_{j}} v_{t}^{w}(t)\delta t + \int\int_{t\in [t_{i}, t_{j}]} (q^{w}_{b}(t)a^{b}(t) - g^{w})\delta t^{2}
   \end{equation}
   \begin{equation}
      \mathbf{v}^{w}(t_{j}) = \mathbf{v}^{w}(t_{i}) + \int_{t_{i}}^{t_{j}} (\mathbf{q}^{w}_{b}(t)a^{b}(t) - \mathbf{g}^{w})\delta t
   \end{equation}
   \begin{equation}
      \mathbf{q}^{w}_{b}(t_{j}) = \int_{t_{i}}^{t_{j}} \mathbf{q}^{w}_{b}(t)\otimes
         \left[ \begin{array}{c}
            0 \\
            \frac{1}{2}\mathbf{w}^{b}(t)
         \end{array}\right]\delta t .
   \end{equation}
\end{subequations}
Similarly, we also have the discrete version by replacing $t_{i}, t_{j}$ with $k, k+1$ in proper way,
\begin{subequations}
   \begin{equation}
      \mathbf{p}^{w}[k+1] = \mathbf{p}^{w}[k] + \mathbf{v}^{w}[k]\Delta t + \frac{1}{2} \mathbf{a}^{w}\Delta t^{2}
   \end{equation}
   \begin{equation}
      \mathbf{v}^{w}[k+1] = \mathbf{v}^{w}[k] + \mathbf{a}^{w}\Delta t
   \end{equation}
   \begin{equation}
      \mathbf{q}^{w}_{b}[k+1] = \mathbf{q}^{w}_{b}[k]\otimes
         \left[ \begin{array}{c}
            0 \\
            \frac{1}{2}\mathbf{w}^{b}\Delta t
         \end{array}\right]
   \end{equation}
\end{subequations}
where
\begin{subequations}
   \begin{equation}
      \mathbf{a}^{w} = \mathbf{q}^{w}_{b}[k](\mathbf{a}^{b}[k] - \mathbf{b}^{a}[k]) - \mathbf{g}^{w}
   \end{equation}
   \begin{equation}
      \mathbf{w}^{b} = \mathbf{w}^{b}[k] - \mathbf{b}^{g}[k]
   \end{equation}
\end{subequations}
or Runge-Kutta approximation, which takes the average between current state and last state
\begin{subequations}
   \begin{equation}
   \begin{split}
	   \mathbf{a}^{w} = & \frac{1}{2}[\mathbf{q}^{w}_{b}[k](\mathbf{a}^{b}[k] - \mathbf{b}^{a}[k]) - \mathbf{g}^{w} \\
	   &+ \mathbf{q}^{w}_{b}[k+1](\mathbf{a}^{b}[k+1] - \mathbf{b}^{a}[k+1]) - \mathbf{g}^{w}]
   \end{split}
   \end{equation}
   \begin{equation}
      \mathbf{w}^{b} = \frac{1}{2}[\mathbf{w}^{b}[k] - \mathbf{b}^{g}[k] + \mathbf{w}^{b}[k] - \mathbf{b}^{g}[k]]
   \end{equation}
\end{subequations}

\subsection{Real Data Analysis}
\label{ssec: real_data_analysis}

When the robot is static, its acceleration should also be stable at local gravity. When the tracked robot moves on a plane, the acceleration begins to change within an acceptable interval. However, the impulses of the acceleration may occur if the robot suddenly changes its state, such as falling down from a step, which might result in large drift in localization due to the double integration of the acceleration. Fig.~\ref{IMU::ori} displays the real IMU data during run-time, which shows that the RealSense IMU has noise with larger  covariance and more unexpected impulses than the Xsens IMU data. When the robot moves suddenly, such as falling down, peaks representing big acceleration occur in the IMU measurements (see Fig.~\ref{xsens}), which are the normal measurements. However, the IMU measurements with larger noise from RealSense may contain the unexpected impulse (see Fig.~\ref{realsense}), which are the fault we want to mitigate. 

In a visual-inertial system, we consider two possible effects of influence of the IMU on the pose estimation:

%When the impulse of acceleration measurement comes up, the pose of robot will drift a lot due the twice intergral of acceleration.

%TODO insert some pic of IMU plot data during robot moving.
\begin{figure*}[tb]
	\centering
	\subfigure[Xsens IMU Measurement Sample]{
		\includegraphics[width=0.45\linewidth]{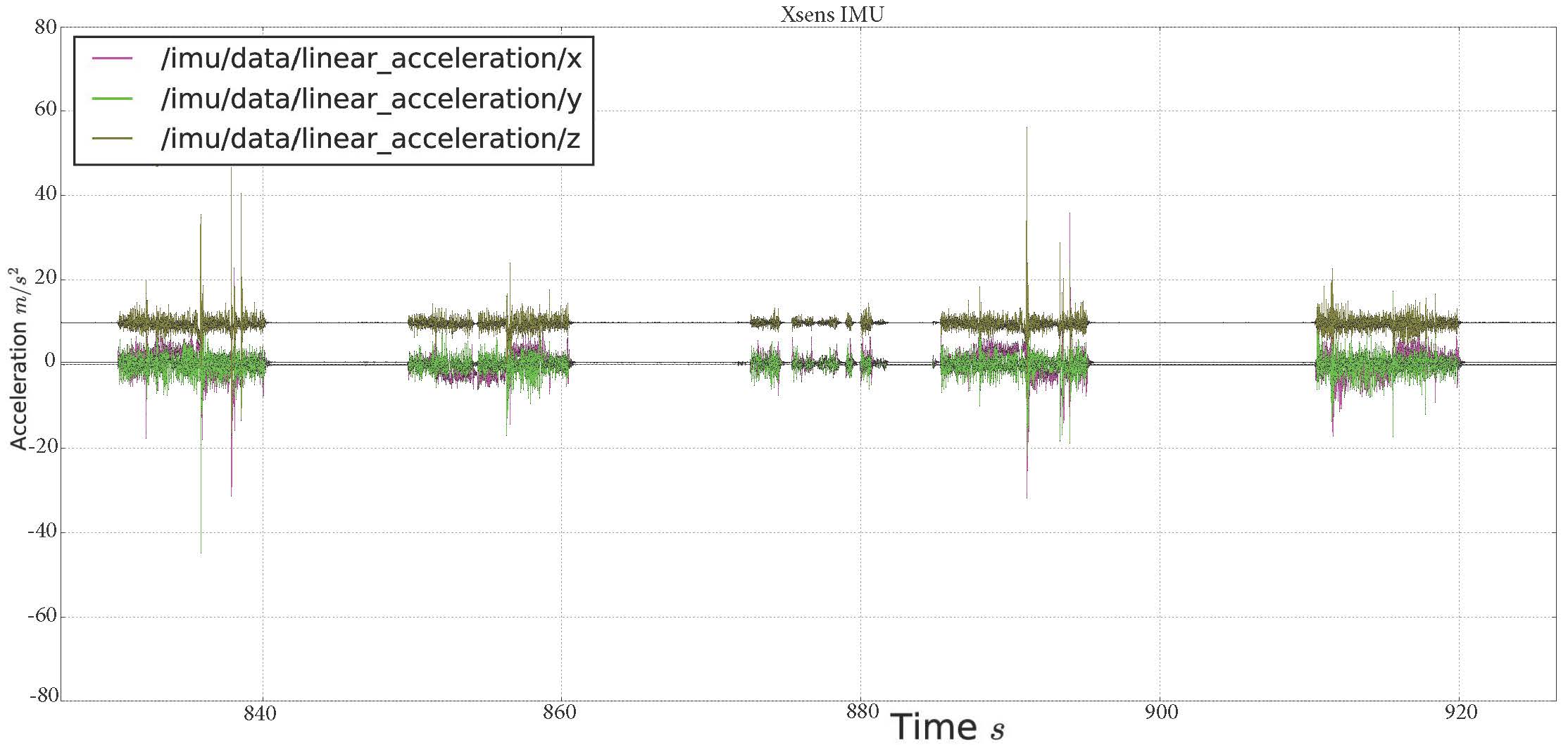}
		\label{xsens}
	}
	\quad
	\subfigure[RealSense IMU Measurement Sample]{
		\includegraphics[width=0.45\linewidth]{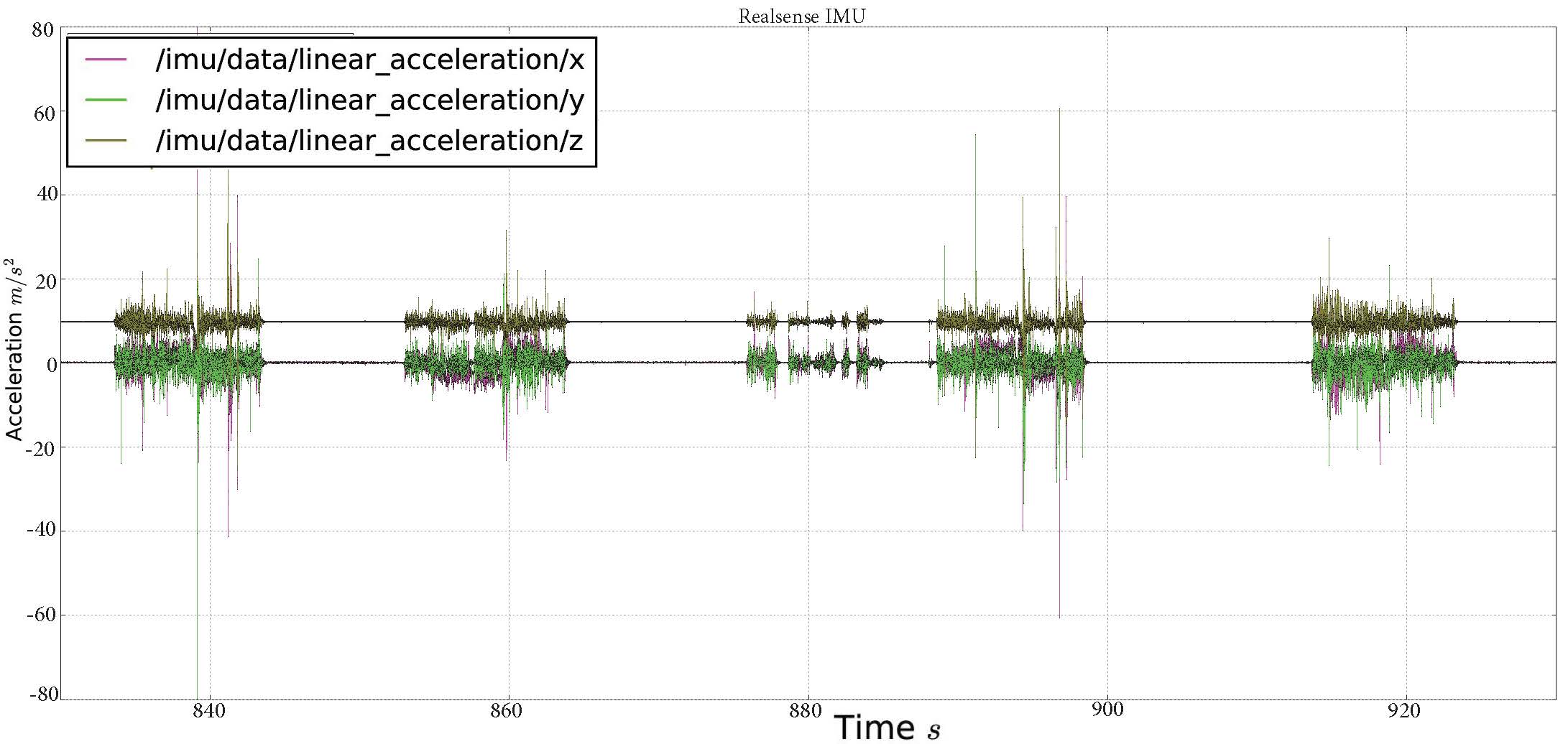}
		\label{realsense}
	}
	\caption{The run-time IMU measurements}
	\label{IMU::ori}
\end{figure*}

%PA1:
%1) Due to the unbalanced IMU sample, the result pose and velocity are affected. For example, the robot starts to move and stops at certain point. The start and the end velocity should be zero, since the integral of acceleration through time are zero. However, it might be non-zero during run-time. Then this drift will accumulate to unpredictable large.

%1) Small errors in acceleration accumulate in the velocity due to the integration, such that it might happen that after some driving around, a constant velocity is estimated, even though the robot is static. This will then lead to a large drift in position.

1) When the robot is abruptly accelerating, for example due to bumping or falling, the IMU measurements may not correctly reflect the changes according to the Nyquist sampling theory, which leads to errors on velocity and pose. For example, the robot starts to move and stop at certain point. The start and the end velocity should be zero, since the integral of acceleration through time is zero. However, the estimated value might be non-zero during run-time. Then the errors in acceleration accumulate in the velocity due to the integration, resulting in unpredictable estimation.

%YIJUN: the bias in the right bottom is not consistent to [-35.-30]
% \begin{figure}[bt]
%    \begin{center}
%       \includegraphics[width = 0.45\textwidth]{pic/velo.jpeg}
%       \caption{The result velocity data, the drift gets larger and larger at time [-35, -30]}
%       \label{velo::ori}
%    \end{center}
% \end{figure}

%PA2:
2) We find that instantaneous accelerations are highly related to the robot instantaneous rotations. Based on the formulation, if the instantaneous rotation is estimated with small errors, the acceleration orientation will be wrong and that error will accumulate in the velocity and position.

It should be noticed that the performance of the visual estimation is not discussed in this part. On one hand, if the vision-based pose estimation is accurate, it will also be influenced by the noise of IMU as constraints. On the other hand, if the visual estimation fails, the IMU data will be dominant in the visual-inertial system.

\section{Methodology}
\label{sec:method}
\vspace{-0.1cm}

For the inertial-based localization, the acceleration measurements will produce significant impulses if the rescue robot falls down or moves rapidly. Some of the impulses lead the robot localization to fail, especially also because it typically goes together with a large visual viewpoint change and blurred images, both of which tend to let the visual odometry fail. To make the inertial-based system more robust and accurate, we divide the approaches into two steps: fault detection and mitigation.

\subsection{Fault Detection}
% When the robot suddenly moves, there are peaks in IMU measurements, which do not reduce the performance of localizaton. However, the low-cost IMU might have many unexpected IMU impulse shown in Fig.~\ref{IMU::ori}, which some impulse could be extremely large. These impulse will have very bad influence on localization.

As mentioned in Sec.~\ref{ssec: real_data_analysis}, we want to detect the unexpected impulse of IMU measurements. 
Following are \textit{Threshold}-based and DTW-based methods to detect these unexpected IMU impulses.

1) \textit{Threshold}: One straightforward method is the \textit{Threshold}-based approach, which identifies any measurement beyond the specified threshold. However, the threshold tuning is tricky: it should not affect any normal measurement and be able to identify as many as possible abnormal measurements. For example, the acceleration value when the robot begins to move suddenly might be similar to that when a glitch occurs.

2) \textit{Dynamic Time Warping~\citep{berndt1994using}}: The DTW algorithm is able to find the optimal match between two time series. Furthermore, it could be used as one kind of metric to measure the similarity of two time series. The characteristic of DTW makes it suitable for our qualification. When we apply DTW to detect the abnormal IMU measurement, the stream of the IMU measurements are grouped into slices of $N$ non-overlapping samples. Our approach is then detecting possible glitches slice-wise. 

The detection is made by comparing the slices (time series) of IMU readings against templates of normal IMU readings. Those templates are previously collected and contain representative examples of acceptable motion patterns of the robot, e.g. driving on smooth ground with different control input.

The measured IMU slices are compared against all templates. If the distance of the best fit is below a certain threshold that measurement slice is classified as normal. Otherwise that slice is considered abnormal.

The implementation details of DTW used in this work can be found in Algorithm~\ref{dtw}. It is with two slices of measurement $\mathbf P \in \mathbb R^{N\times d}$ and $\mathbf Q \in \mathbb R^{M\times d}$. We use the concatenation of signals to represent the slice. $N$, $M$ are the discretized lengths and $d$ is the dimension of original measurements for comparison. The dimension $d$ is typically 3 for a 3-axis accelerometer or 6 when also taking a 3-axis gyroscope into account. 

\begin{algorithm}[t]
	\caption{DTW}
	\label{dtw}
	\hspace*{\algorithmicindent} \textbf{Input} Two slice $\mathbf P \in \mathbb R^{N\times d}$ and $\mathbf Q \in \mathbb R^{M\times d}$. \\
	\hspace*{\algorithmicindent} \textbf{Output} distance
	\begin{algorithmic}[1]
		\State $T\gets array(M+1)$ \Comment{index from 0 to M}
		\State $T[0] \gets 0$
		\State $T[i] \gets \infty$ for $i \in \{1,\cdots,M\}$.
		\For{$level  \in \{1,\dots,N\}$}
		\State$ul\_corner\gets T[0]$
		\State$T[0]\gets \infty$
		\For{$i  \in \{1,\dots,M\}$}
		\State$min\_v\gets dist(\mathbf p_{level}, \mathbf q_i) + \min (ul\_corner,T[i-1],T[i])$
		\State$ul\_corner\gets T[i]$
		\State$T[i]\gets min\_v$
		\EndFor
		\EndFor
		\State $distance\gets T[M]$
	\end{algorithmic}
\end{algorithm}

The basic distance function for two vectors $\mathbf p_i\in \mathbb{R}^{d}$, $\mathbf q_j\in\mathbb{R}^{d}$, which are the $i$-th and $j$-th rows of $\mathbf P$ and $\mathbf Q$, respectively, is
\begin{equation}
\label{eq:dist}
dist(\mathbf p_i, \mathbf q_j) = || \mathbf p_i- \mathbf q_j || ^2.
\end{equation}

To detect the abnormal measurement, multiple templates are required, because different measurement patterns may vary. The computational efficiency of the detection algorithm is also important, due to the high frequency of the IMU measurements. Assuming $k$ templates with length $N$ and IMU measurements with length $M$, and $M \ge N$ for convenience, then the time and space complexity of detection for each test are $\mathcal O(kM^2)$ and $\mathcal O(M)$, respectively. Moreover, parallel computing  is utilized to increase the speed of the detection, since comparing with each template is independent. In this case, the time complexity is $\mathcal O(\frac{k}{l}M^2)$ ($l$ is the number of parallel threads), and the space complexity increases to $\mathcal O(lM)$.

% In addition, with a high frequency of IMU measurement, if detection algorithm is equipped, it has to be with low cost. Assume $k$ templates with length $N$ each are used and $M<=N$ for convenience. Then the time complexity and space complexity of detection for each test are $\mathcal O(kN^2)$ and $\mathcal O(N)$ with a sequentially run.
% However, since comparing with each template is independent to one another, parallel computing is available with time complexity $\mathcal O(\frac{k}{l}N^2)$ ($l$ is the number of threads in parallel), and a larger space complexity $\mathcal O(kN)$. Thus multiple threads are used to boost the detection. In our implementation, it is fast enough to keep pace with IMU measurement.

\vspace{-0.1cm}
\subsection{Fault Mitigation}

After detecting abnormal IMU measurements we need to replace those samples with other data, which will hopefully lead to better visual-intertial localization results. We have implemented the following strategies for the two detection methods:

%For the aspect of measurement source, we can find out this extremely high impulse from acceleration measurement and replace it with normal data, which leads the localization work properly. This strategy can be applied to improve the inertial system after fault detection. Specifically, the details of the strategy that performs on the two detection methods are described as follows:

1) \textit{Threshold}: After the fault is detected by a specific threshold, we can replace the measurement with either the threshold or the average of the last $n$ samples. The method inhibits the abnormal measurements, but it may also break the balance of the original measurements.

2) \textit{Dynamic Time Warping}: In contrast to replacing the single measurement in the \textit{Threshold}-based method, the DTW-based method exploits a better way that combines the replaced values with reasonable close previous measurements. As discussed previously, we collect data templates from the robot at different states for fault detection. If a fault is detected, the complete slice of IMU measurements will be replaced by the best fitting template. We think that this better keeps the balance of the acceleration measurements.
The templates represent series of the normal IMU measurements. It is reasonable, because once abnormal measurements appear, a large DTW distance arises on the different templates. Since the captured measurement slice still contains useful data beside the abnormal, the closest template is still a good choice on the pattern distribution.

\subsection{Integration to VINS Framework}

\begin{figure}
	\begin{tikzpicture}%[background rectangle/.style={fill=blue!45, line width=1mm}, show background rectangle]
	\node[module] (imu) at (0, 0) {IMU \\Measurement};
	\node[module, fill=blue!60] (dtw) at (3.2, 1) {DTW};
   \node[module, fill=red!80] (th) at (3.2, -1) {Threshold};

   \node[module, fill=blue!60] (tmp) at (3.2, 2.2) {Templates};

   \node[module] (vins) at (6.4, 0) {VINS-Mono};

   % \node[module] (pos) at (6, 1.5) {Pose};

   \draw[->] (imu) |- (tmp);
   \draw[->, dashed, line width=0.4mm] (imu) -- (vins);
   \draw[->, line width=0.4mm] (imu) |- (th);
   \draw[->, line width=0.4mm] (imu) |- (dtw);
   \draw[->, line width=0.4mm] (tmp) to (dtw);
   \draw[->, line width=0.4mm] (th) -| (vins);
   \draw[->, line width=0.4mm] (dtw) -| (vins);

   \begin{pgfonlayer}{background}
      \node [fill=olive!30, double, draw, line width=0.3mm, fit= (vins) (th) (dtw)  (tmp) (imu)] {};
      % \node [fill=olive!50, minimum width=40mm, draw, line width=0.5mm, fit=  (dtw)  (tmp) ] {};
      % \node [fill=red!30, double, draw, line width=0.3mm, fit= (th)] {};
   \end{pgfonlayer}
	\end{tikzpicture}
	\caption{Overview of the three versions of our improved VINS-Mono framework: red: Threshold version; blue: DTW-based approach; dashed line: original VINS-mono.}
	\label{fig:overview}
\end{figure}
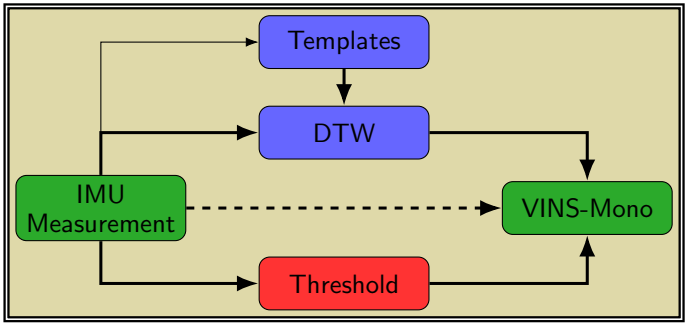

\begin{figure}[tb]
	\centering
	\subfigure[X-Y side view]{
		\includegraphics[width=0.4\linewidth]{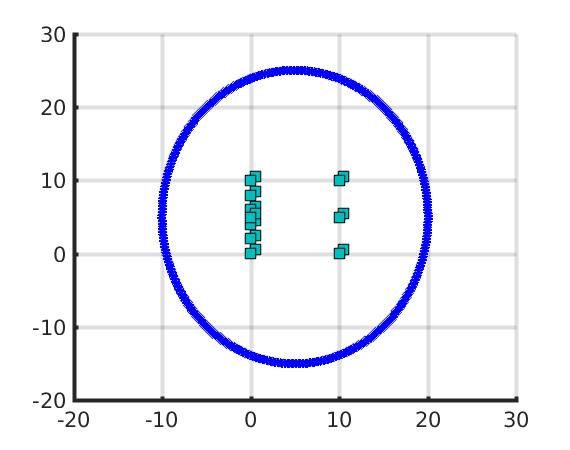}
	}
	\subfigure[X-Z side view]{
		\includegraphics[width=0.4\linewidth]{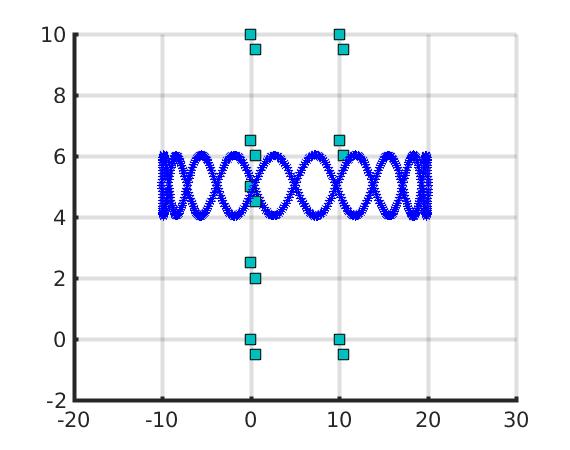}
	}
	\subfigure[Y-Z side view]{
		\includegraphics[width=0.4\linewidth]{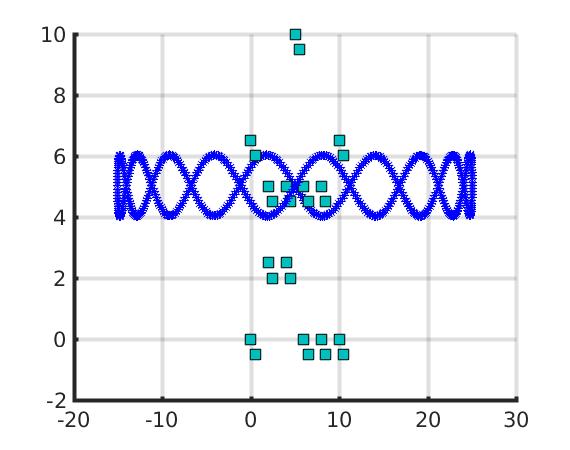}
	}
	\subfigure[Brief view]{
		\includegraphics[width=0.4\linewidth]{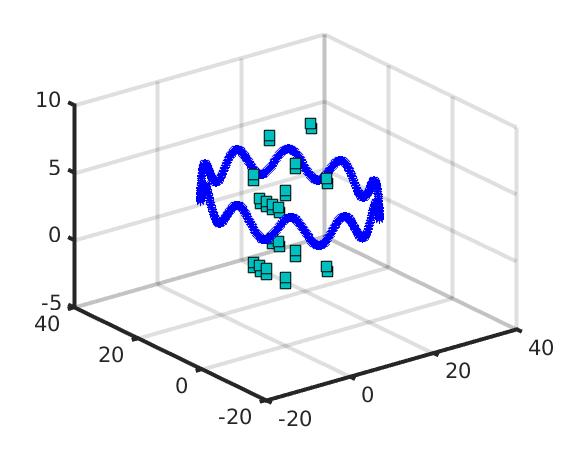}
	}
	\caption{Simulated trajectories and landmarks}
	\label{fig:sim_sample}
\end{figure}

\begin{figure}[tb]
\centering
\includegraphics[width=0.8\linewidth]{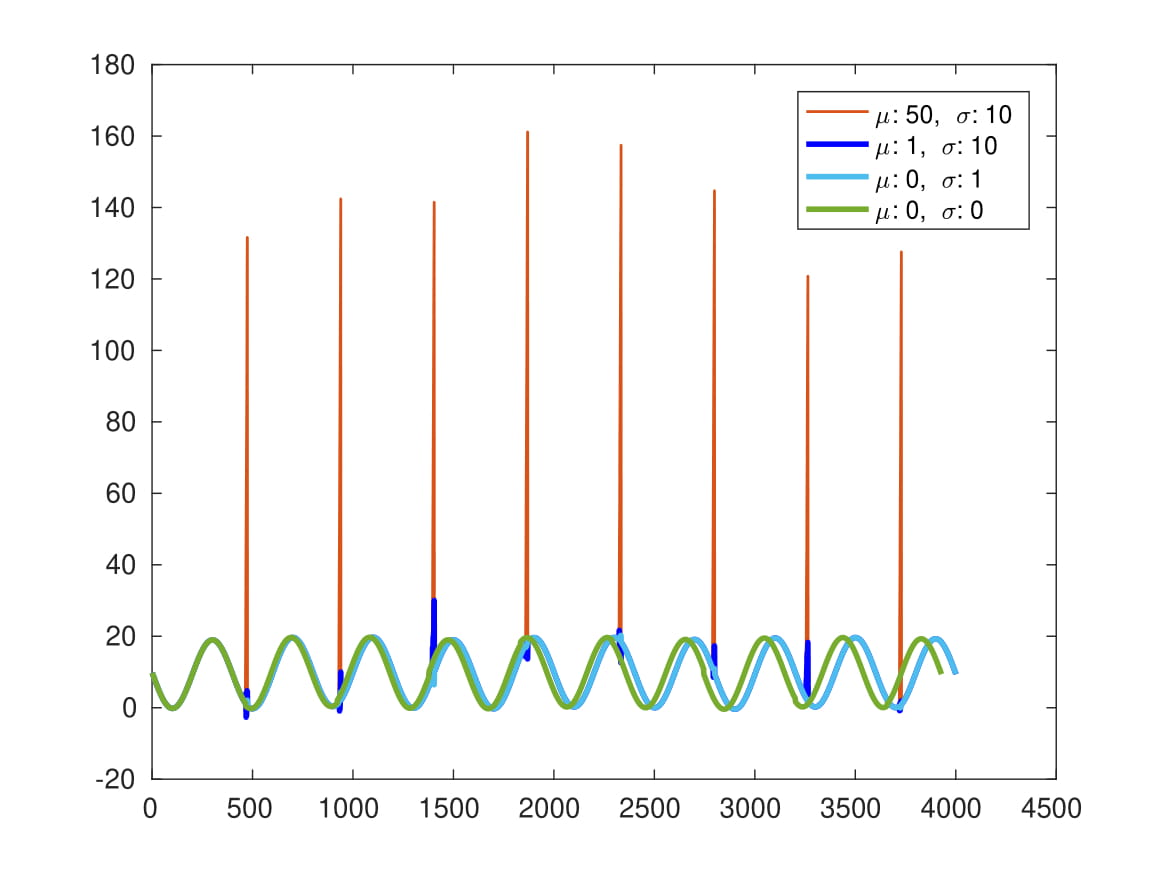}
\caption{Noise on acceleration along $z$-axis}
\label{fig:noise_on_acc_z}
\end{figure}

To evaluate the performance of the two different fault mitigation methods, we embedded the methods into the VINS-Mono \citep{qin2017vins} system. Fig.~\ref{fig:overview} demonstrates the improved system with the IMU abnormality mitigation mechanism. Instead of feeding IMU measurements directly into VINS-Mono (dashed line), fault detection and mitigation are applied beforehand. The proposed \textit{Threshold}-based (red part) and DTW-based (blue part) methods are applied separately on IMU measurements. 
For the \textit{Threshold}-based method, we replace the abnormal values with the threshold when the abnormal measurements are detected.
For the DTW-based approach, we first try to match the IMU measurements with all pre-obtained templates. If none is well-matched, it means the measurements might be the glitch error which we want to mitigate. If that is the case, the best fitting template is used as replacement. Then the corresponding replacement takes over the original data, leading localization into better performance.

\begin{figure*}[t]
	\centering
	\subfigure[$\mathcal{N}(0, 1)$ top view]{
		\includegraphics[width=0.3\linewidth]{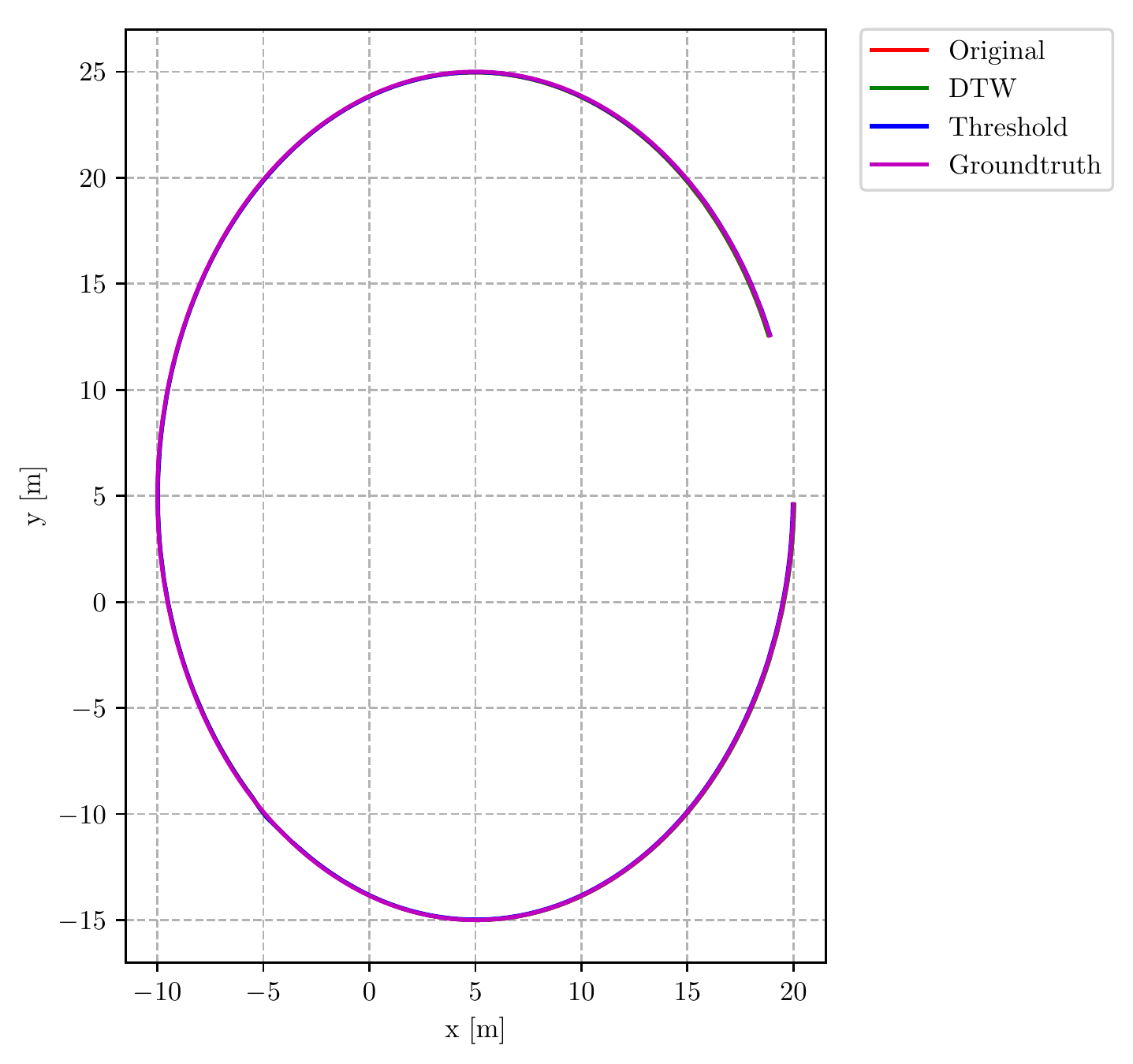}
	}
	\subfigure[$\mathcal{N}(1, 10)$ top view]{
		\includegraphics[width=0.3\linewidth]{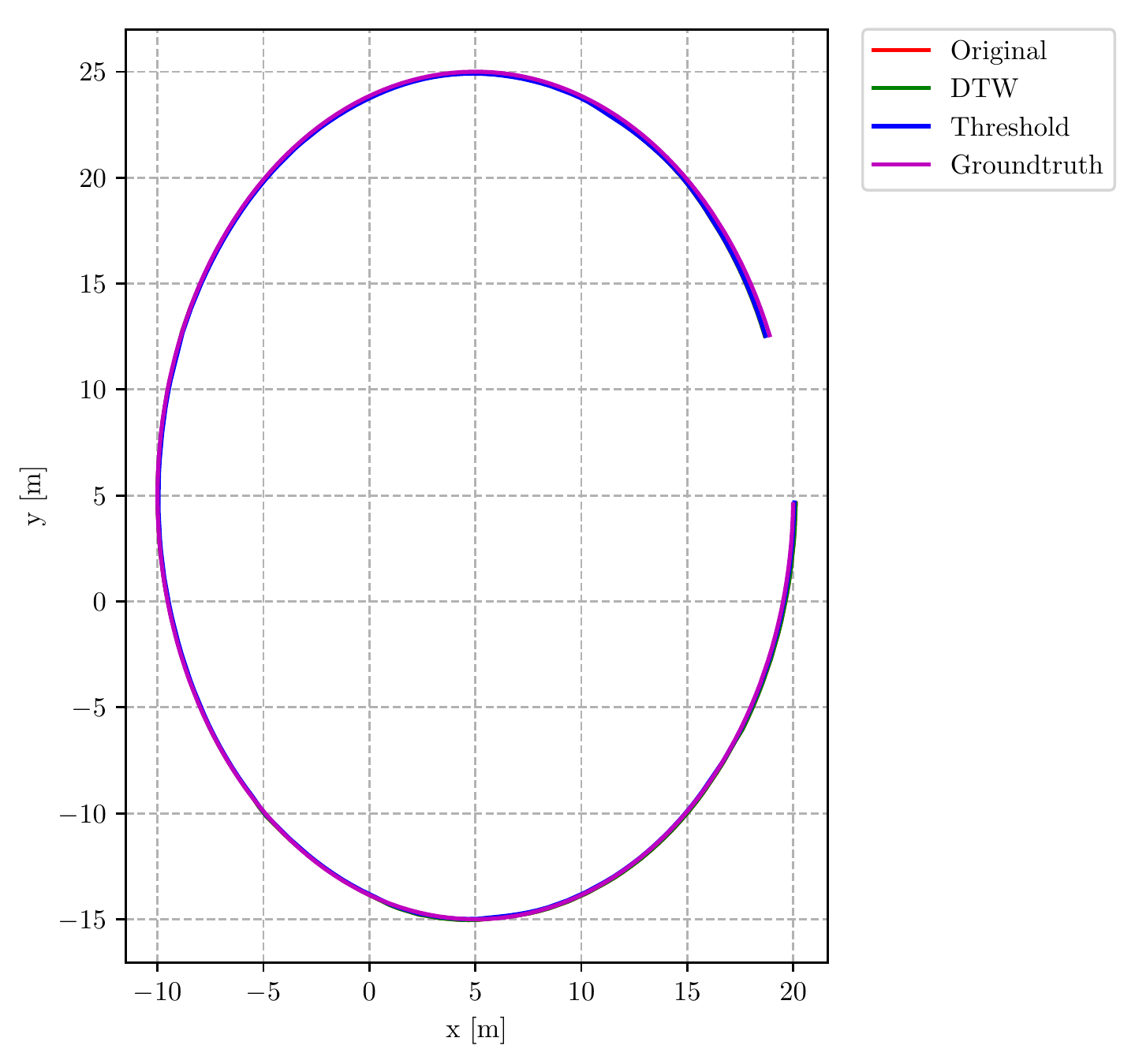}
	}
	\subfigure[$\mathcal{N}(50, 10)$ top view]{
		\includegraphics[width=0.3\linewidth]{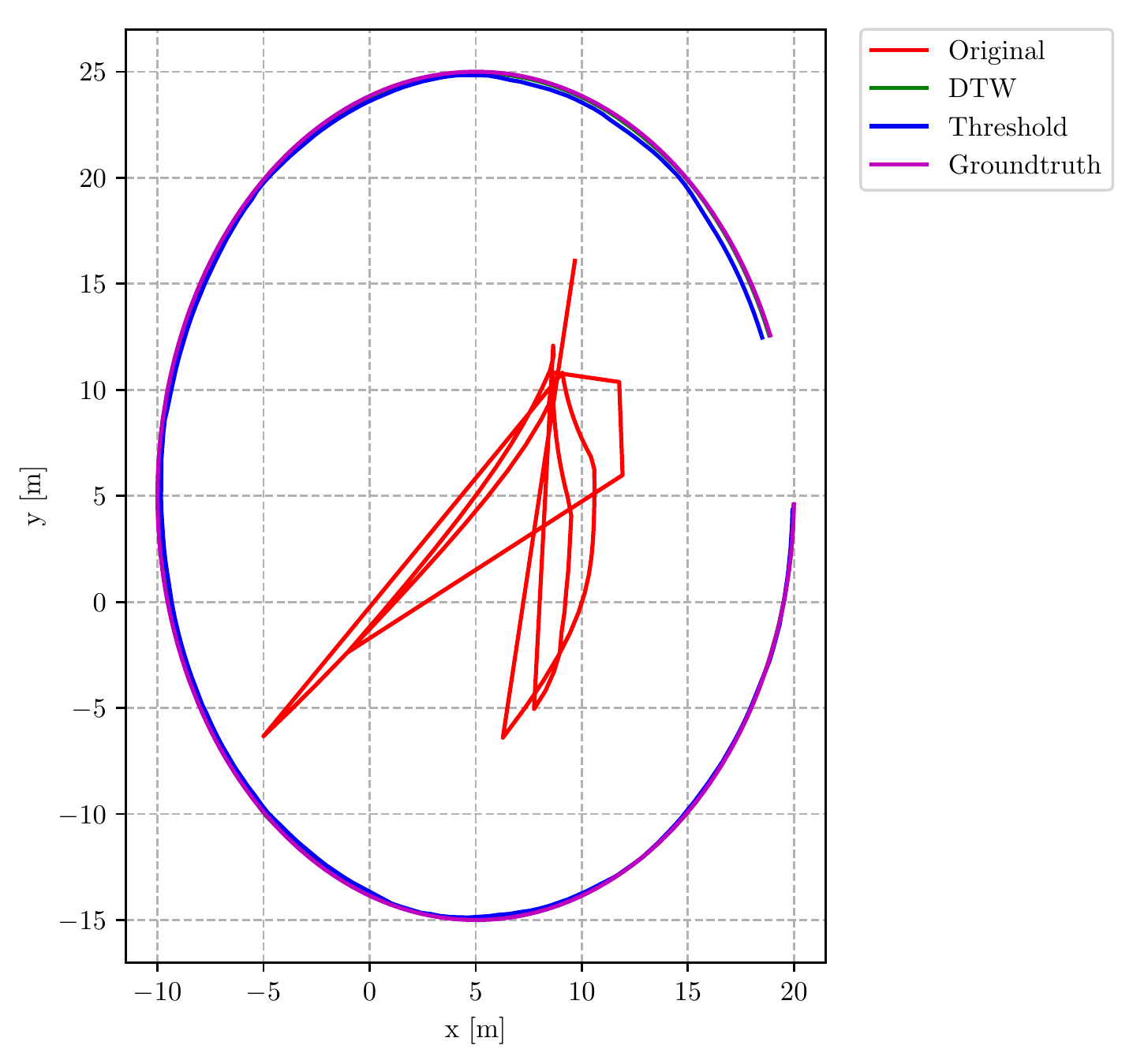}
	}
	\subfigure[$\mathcal{N}(0, 1)$ side view]{
		\includegraphics[width=0.3\linewidth]{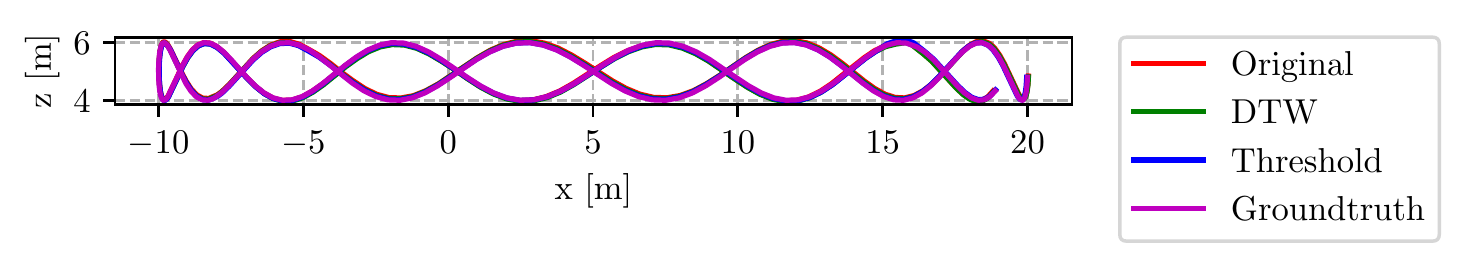}
	}
	\subfigure[$\mathcal{N}(1, 10)$ side view]{
		\includegraphics[width=0.3\linewidth]{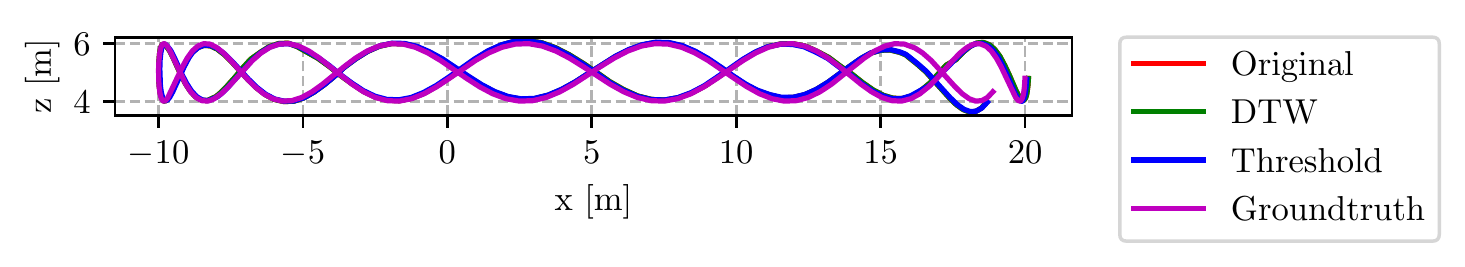}
	}
	\subfigure[$\mathcal{N}(50, 10)$ side view]{
		\includegraphics[width=0.3\linewidth]{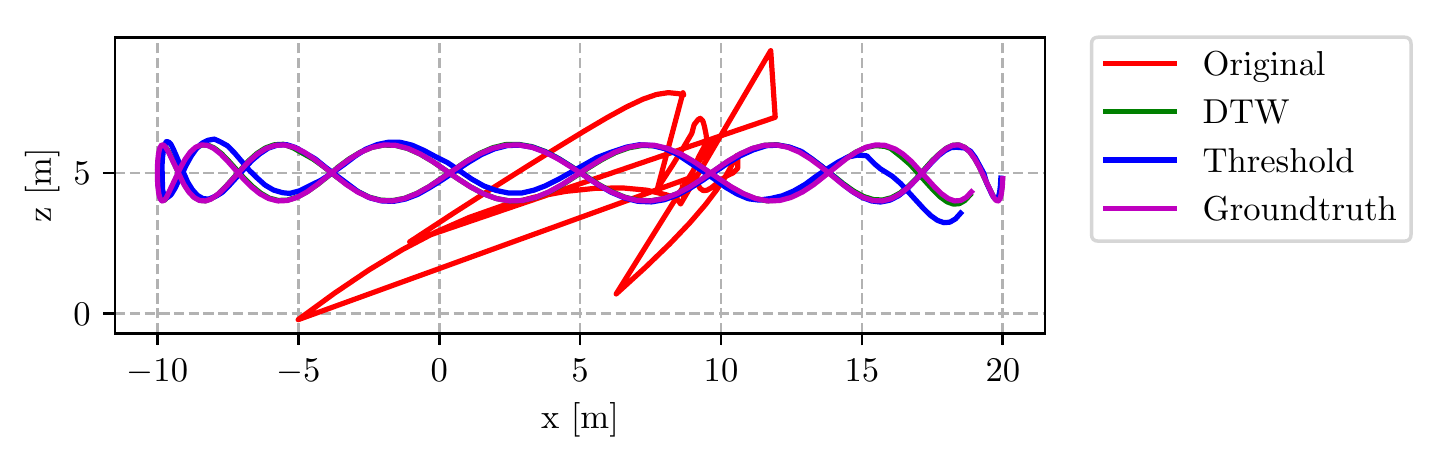}
	}
	\subfigure[$\mathcal{N}(0, 1)$ translation error]{
		\includegraphics[width=0.3\linewidth]{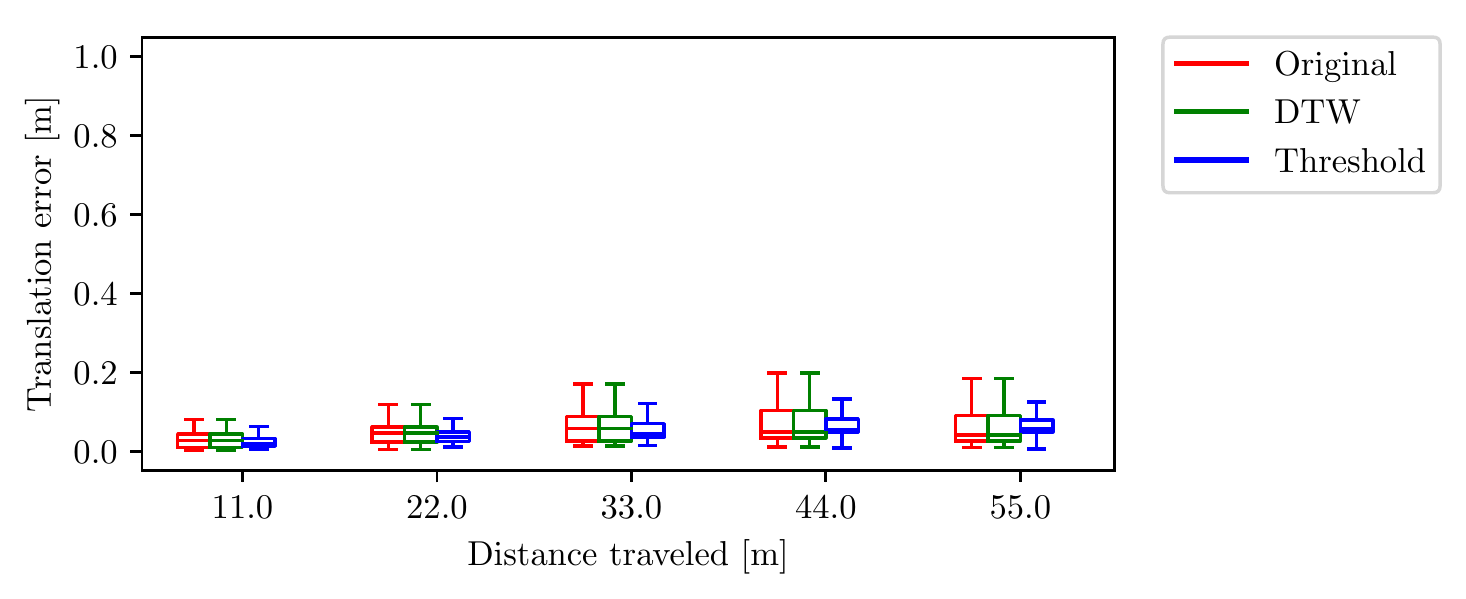}
	}
	\subfigure[$\mathcal{N}(1, 10)$ translation error]{
		\includegraphics[width=0.3\linewidth]{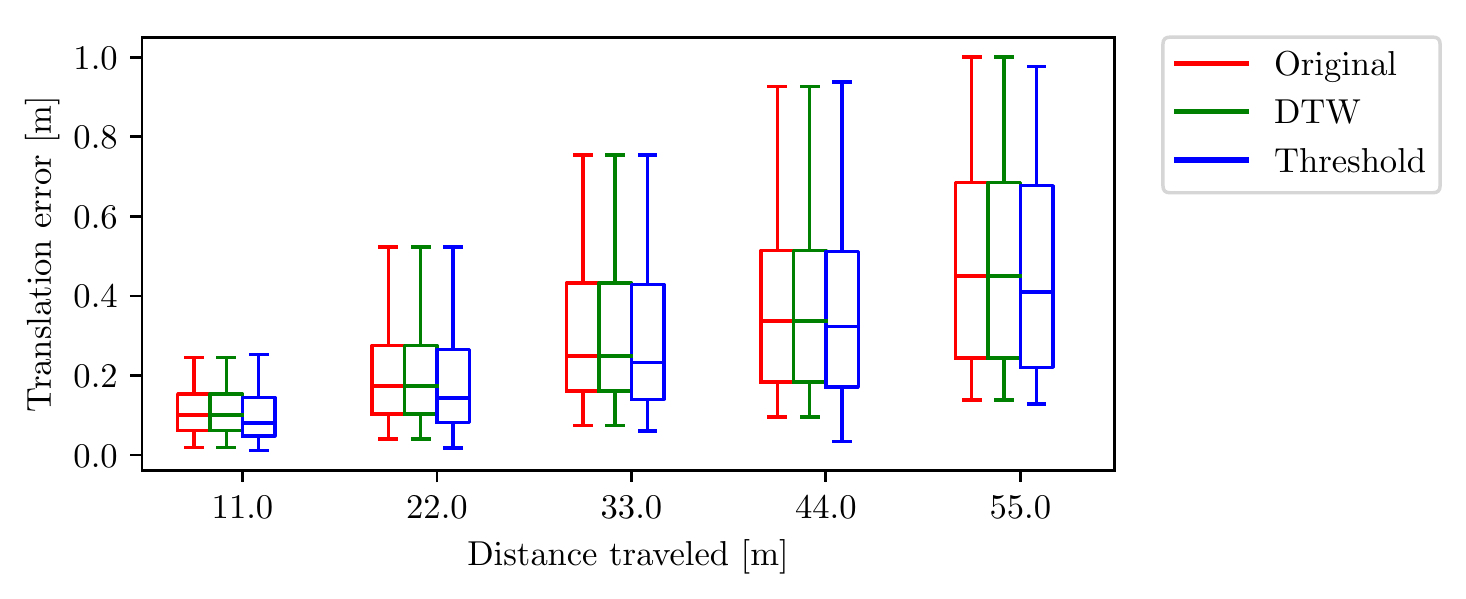}
	}
	\subfigure[$\mathcal{N}(50, 10)$ translation error]{
		\includegraphics[width=0.3\linewidth]{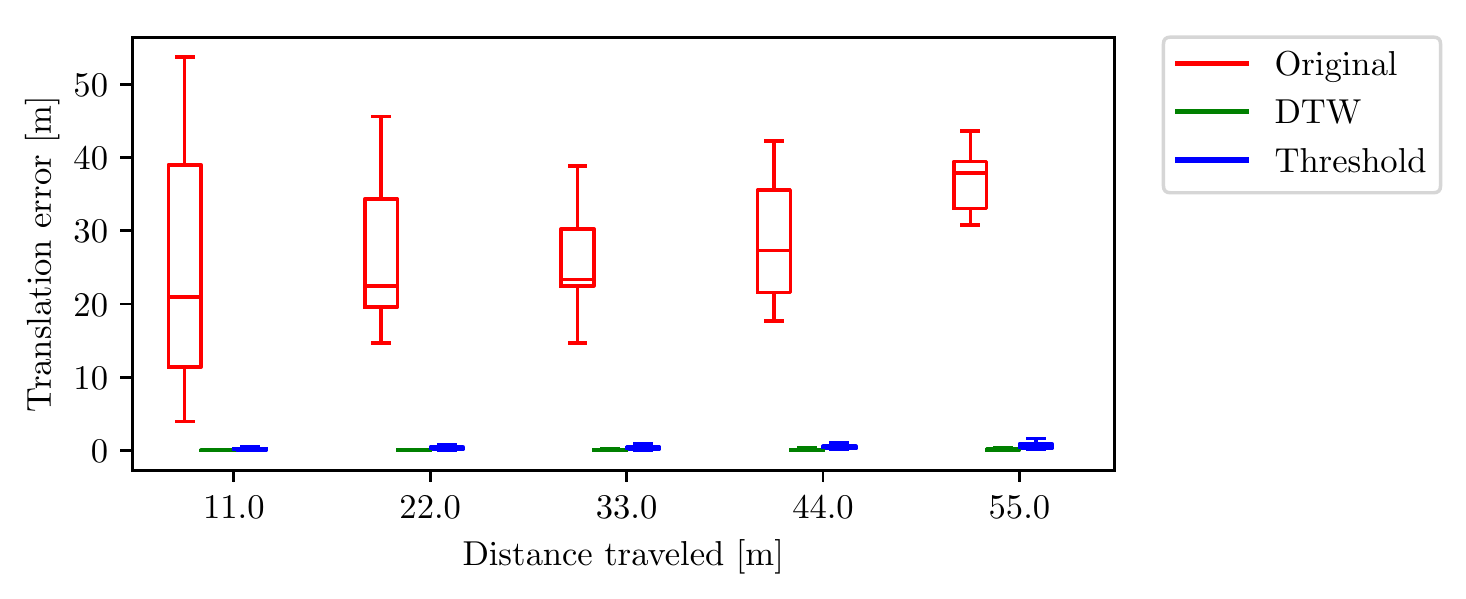}
	}
	\caption{Improvement on localization with different methods and glitch errors}
	\label{fig:sim_exp}
\end{figure*}

\section{Experiments and Analysis}
\label{sec:exp}
In this section, we compare the performance of the \textit{Threshold}-based and DTW-based framework with original VINS-Mono on both simulation and real robot data. All the pose estimation accuracy in the experiments is evaluated by \texttt{rpg\_trajectory\_evaluation}\footnote{\url{https://github.com/uzh-rpg/rpg_trajectory_evaluation}} \citep{Zhang18iros}.

%For the convenience of parallel computing on distance, templates are with same length from origin.

%YIJUN: use multi-thread or single thread.
% \subsubsection{Select Template}
% \textcolor{red}{Here shows how to select the templates, In Fault Isolation, I write the template can reflect the property of original distribution of patterns.}

%YIJUN: have a subsection{Setting} to tell the detailed parameter in threshold and DTW

%YIJUN: add the experiment time cost for DTW and tell it can run in real time.

%In order to illustrate the effectiveness of DTW, we compare with the threshold-based method, which replaces the acceleration data when the measurement is greater than predefined threshold.

\subsection{Simulation}

To exclude the influence of localization based on vision, we perform simulation experiments with the same number of feature points in each view. Thus we can concentrate on the influence of the IMU to the visual-inertial framework. In this experiment, the number of templates is $k=10$ and they have a length of $N = 10$, the length of the measured slices is $M = 40$ and the dimension for the 3-axis accelerometer with 3-axis rate gyroscope is $d=6$. The simulated IMU and keypoints data are generated from given trajectories and 3D landmarks, where 2000 poses and 36 landmarks are used in this simulation (see Fig.~\ref{fig:sim_sample}).

To make this data closer to real data, we add Gaussian noise to the idealistic raw data. In addition, to simulate the glitches, we add noise with bigger mean $\mu$ (offset) and covariance $\sigma$ to a few samples. As shown in Fig.~\ref{fig:noise_on_acc_z}, the acceleration along the $z$-axis is influenced by noise with different $\mu$ and $\sigma$ and the green line represents acceleration along $z$-axis with no noise.

Then we run the improved VINS-Mono system on these datasets with different amounts of glitch to compare the two fault mitigation approaches.
Fig.~\ref{fig:sim_exp} shows the results of boosting localization accuracy with two different methods and three different glitch settings. It is found that the original VINS-Mono (red line) works well when the glitch error is small ($\mathcal{N}(0,1)$) but its ability on pose estimation decreases as the noise gets bigger.
It is obvious that the original VINS-Mono fails to localize when the glitch is too large, where \textit{Threshold}-based and DTW-based methods improve the estimation results. In addition, Table~\ref{tab:rmse_trans} gives a quantitative comparison, which shows that the DTW-based method performs best.
Tables~\ref{tab:overall_trans} and~\ref{tab:overall_yaw} display the overall relative translation and yaw error based on the three glitch settings, which indicates that both methods improve the performance of the original VINS-Mono system. It can also be observed that the DTW-based method yields better results than the \textit{Threshold}-based one.

\begin{table}[t!]
	\centering
	\caption{Root mean square of translation on simulation data}
	\begin{tabular}{l||l|l|l}
		\hline\hline
		      &      noise\_0\_1 & noise\_1\_10 &  noise\_50\_10 \\
		      \hline
		      trunc &     0.04 & 0.14 &  0.32 \\
		      \hline
		      dtw &     0.06 & 0.15 &  0.06 \\
		      \hline
		      original &     0.06 & 0.15 &  18.05 \\
		\hline
	\end{tabular}
	\label{tab:rmse_trans}
\end{table}

\begin{table}[t]
	\centering
	\caption{Overall relative translation error on simulation data}
	\begin{tabular}{l||l|l|l}
		\hline\hline
	      &      trunc & dtw &  original \\
	      \hline
	      7.00m &     0.16 & 0.07 &  2.50 \\
	      \hline
	      14.00m &     0.26 & 0.10 &  3.03 \\
	      \hline
	      21.00m &     0.37 & 0.14 &  3.54 \\
	      \hline
	      28.00m &     0.43 & 0.17 &  4.02 \\
	      \hline
	      35.00m &     0.51 & 0.20 &  4.29 \\
		\hline
	\end{tabular}
	\label{tab:overall_trans}
\end{table}

\begin{table}[t]
	\centering
	\caption{Overall relative yaw error on simulation data}
	\begin{tabular}{l||l|l|l}
		\hline\hline
		      &      trunc & dtw &  original \\
		      \hline
		      7.00m &     0.61 & 0.08 &  9.47 \\
		      \hline
		      14.00m &     0.49 & 0.08 &  9.66 \\
		      \hline
		      21.00m &     0.53 & 0.09 &  10.96 \\
		      \hline
		      28.00m &     0.51 & 0.10 &  12.81 \\
		      \hline
		      35.00m &     0.55 & 0.10 &  14.71 \\
		\hline
	\end{tabular}
	\label{tab:overall_yaw}
\end{table}

\begin{figure*}[]
\centering
\includegraphics[width=0.75\linewidth]{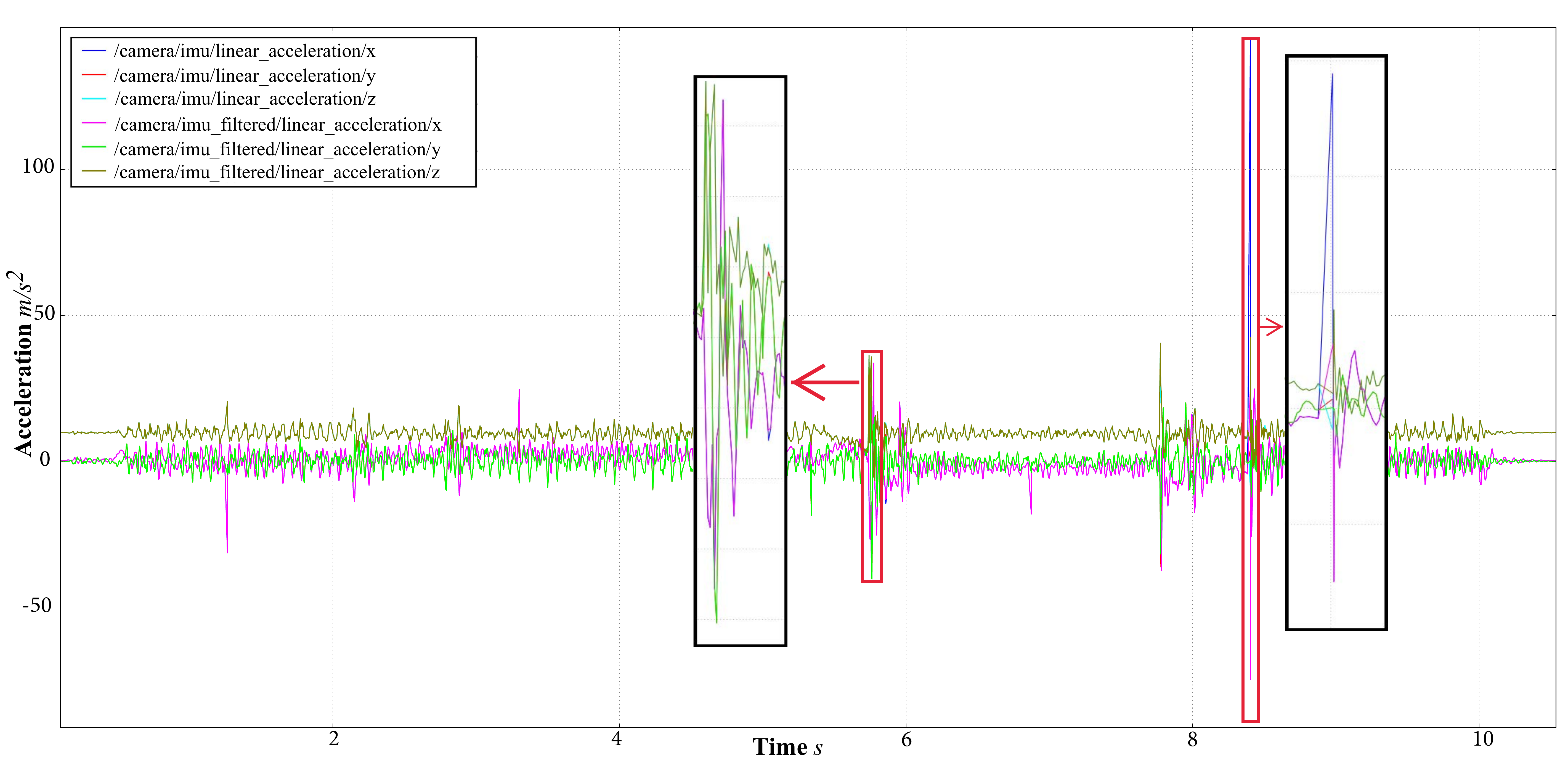}
\caption{A sample of IMU measurement and fault detection and mitigation result by DTW detection method for trial 2.}
\label{fig:imu_sample}
\end{figure*}

\begin{figure*}[tb]
	\centering
	\subfigure[Trial 1 top view]{
		\includegraphics[width=0.4\linewidth]{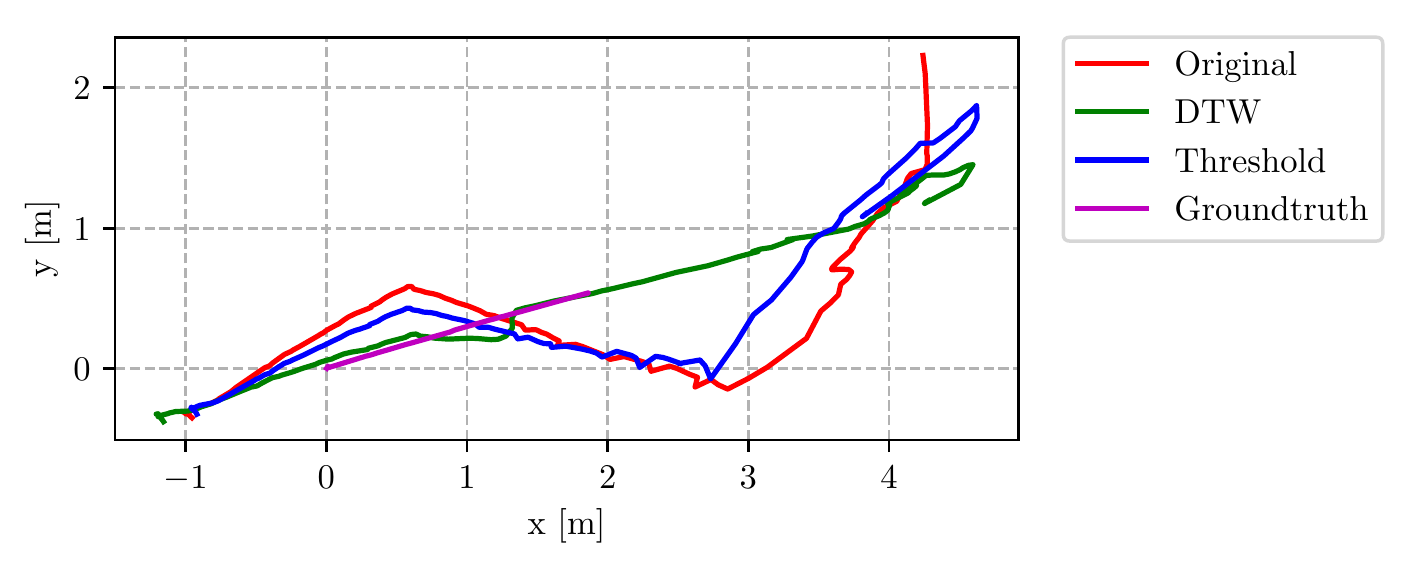}
	}
	\subfigure[Trial 2 top view]{
		\includegraphics[width=0.45\linewidth]{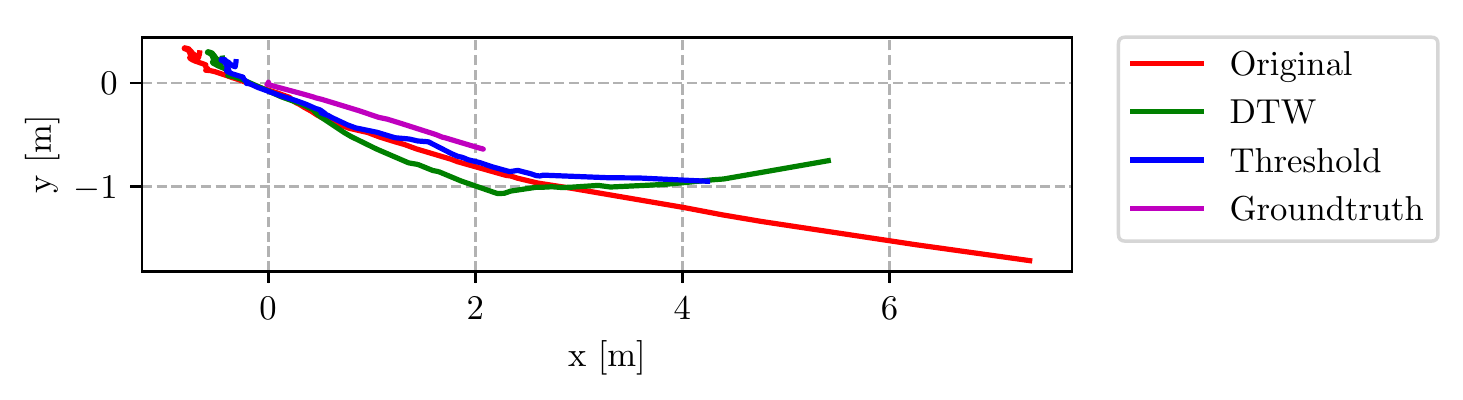}
	}
	\subfigure[Trial 1 side view]{
		\includegraphics[width=0.4\linewidth]{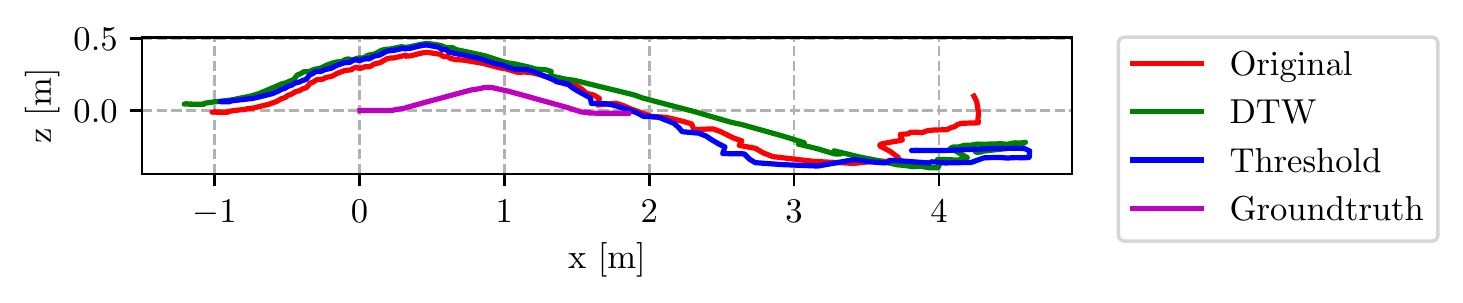}
	}
	\subfigure[Trial 2 side view]{
		\includegraphics[width=0.45\linewidth]{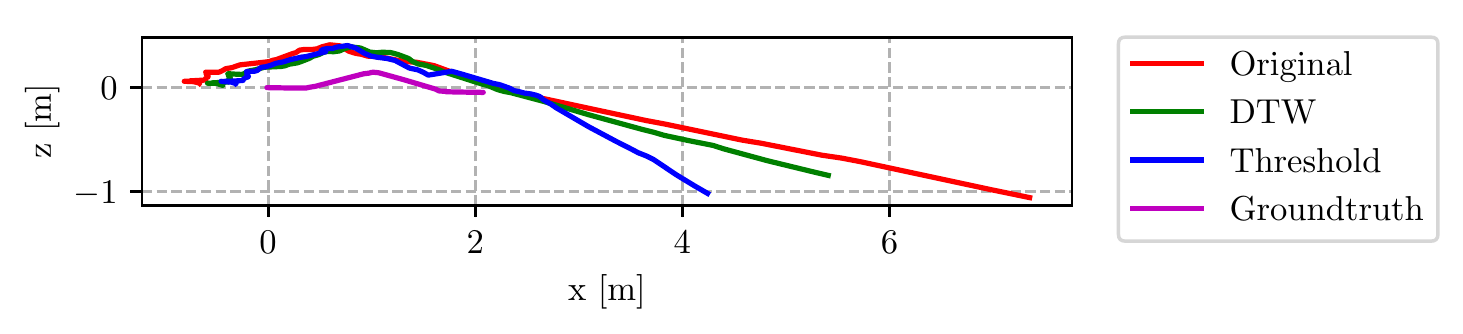}
	}
	\caption{Real robot trials on localization with different methods.}
	\label{fig:real_exp}
\end{figure*}

\subsection{Real Robot Experiment}

Finally, we test our approach also on a real robot. Firstly, we collect IMU data templates from measurements when the robot moves smoothly on different states, including straight moving, circling, climbing and downhill. Then IMU data is captured in the test terrain to compare with templates during detection.

In this experiment, the number of templates is $k=10$ and they have a length of $N = 40$, the length of the measured slices is $M = 40$ and the dimension for the 3-axis accelerometer is $d=3$. 

We do not take the angular velocity into account for two reasons. Firstly, the angular velocity is captured by a gyroscope, which is not largely affected by the robot's acceleration; Secondly, the visual odometry already has trustworthy results on orientation estimation, which is different from the translation of visual odometry, which suffers from an unknown scale factor.
  
The real robot experiments are performed on the small rescue robot standard test scenarios build with wooden ramps. The hardware platform to run VINS-Mono is an Intel NUC (i7-8559U@2.7-4.5GHz and 16GB Memory without GPU) connected to an Intel RealSense D435i camera. Image and IMU data are all captured by this camera. Additionally, the ground truth robot poses are captured with an OptiTrack\footnote{\url{https://www.optitrack.com}}  System. The multi-threaded C++ DTW implementation only takes about $0.6ms$ for each computation with all templates, which is about four times faster than single thread.

When collecting templates for the DTW-based method, we should ensure that no abnormal IMU data is included. For that, these templates can be generated from either high-end IMUs (like Xsens) or the reliable historical measurements of cheap IMUs. In these experiments we use the latter approach.

Fig.~\ref{fig:imu_sample} depicts the IMU measurement and fault detection and mitigation results, which show the efficiency of the proposed method. The \textit{Threshold} detection method has a quite straightforward result, which is not displayed in the paper. The first one within red box is detected when robot fall down from a ramp. The second one within red box is detected when the $x-$acceleration has an extremely high impulse.
%For evaluation, we compare the localization results with the groud truth captured by OptiTrack System.

Fig.~\ref{fig:real_exp} shows the results of the real robot experiments, which indicate that both methods improve the performance. The DTW keeps a better appearance compared to the ground truth, ignoring the scale factor from Trial 1. It should be noted that the groundtruth is shorter than the estimated ones due to the scaling problem instead of missing data. The scale factor can be reduced further when loop closures happen. In the Trial 2, our algorithm prevents the odometry drift a lot on the $x$-axis. As introduced in Sec~\ref{sec:problem_analysis}, the presence of high acceleration impulses will result in unpredictable estimations. The original VINS-Mono has a large continuous drift due to a faulty IMU measurement, which leads to a wrongly estimated large velocity. However, the other two improved methods have no endless drift. Even though there is a big gap between the estimated result and the ground truth, the improved result shows larger probability to be reduced by further optimization.

\vspace{-0.3cm}

\section{Conclusions}
\label{sec:conclusion}
\vspace{-0.3cm}

This paper analyzes the effect of abnormal IMU measurements on visual inertial systems. Glitches in the IMU measurement can make the estimated odometry inaccurate and thus may result in failure of localization, a problem which is especially severe for ground robots. This is because lots of noise is present on the IMU measurements due to the ground contact forces. 
This paper proposes two different IMU preprocessing methods, which are the \textit{threshold}-based and the DTW-based method. 

We demonstrate our methods on both simulated data and real robot experiments. The simulation experiment shows that both methods are very well capable of detecting and mitigating simulated abnormal IMU data and the quantitative data reveals the DTW-based approach as superior to the threshold method. Furthermore, these two methods are able to improve the accuracy of visual-inertial localization.  
The real robot experiments also indicate a good detection of IMU glitches. But the experimental results of the integration into the VINS-Mono framework remain somewhat inconclusive, since the framework exhibits a significant overall scaling problem for all experiments. Moreover, the proposed two methods have limited improvement on localization for rescue robots. 

As future work will further improve the VINS integration to remedy the scaling problems. Firstly, a more flexible template generation will be developed instead of fixed templates, to better represent the terrains information. Secondly, the mitigation strategy will be changed from the templates to data sampling from the distribution of the original IMU measurements. Finally, we will test our method on more difficult terrains and scenarios to show the robustness. 
% \begin{ack}
% Place acknowledgments here.
% \end{ack}
\tiny
\bibliography{ifacconf}             % bib file to produce the bibliography
                                                     % with bibtex (preferred)

%\begin{thebibliography}{xx}  % you can also add the bibliography by hand

%\bibitem[Able(1956)]{Abl:56}
%B.C. Able.
%\newblock Nucleic acid content of microscope.
%\newblock \emph{Nature}, 135:\penalty0 7--9, 1956.

%\bibitem[Able et~al.(1954)Able, Tagg, and Rush]{AbTaRu:54}
%B.C. Able, R.A. Tagg, and M.~Rush.
%\newblock Enzyme-catalyzed cellular transanimations.
%\newblock In A.F. Round, editor, \emph{Advances in Enzymology}, volume~2, pages
%  125--247. Academic Press, New York, 3rd edition, 1954.

%\bibitem[Keohane(1958)]{Keo:58}
%\newblock \emph{Power and Interdependence: World Politics in Transitions}.
%\newblock Little, Brown \& Co., Boston, 1958.

%\bibitem[Powers(1985)]{Pow:85}
%\newblock Is there a way out?
%\newblock \emph{Harpers}, pages 35--47, June 1985.

%\bibitem[Soukhanov(1992)]{Heritage:92}
%\newblock \emph{{The American Heritage. Dictionary of the American Language}}.
%\newblock Houghton Mifflin Company, 1992.

%\end{thebibliography}

% \appendix
% \section{A summary of Latin grammar}    % Each appendix must have a short title.
% \section{Some Latin vocabulary}              % Sections and subsections are supported
                                                                         % in the appendices.
\end{document}